# Adaptive Contrast Enhancement and Optimised Feature Matching for RootSIFT-Based Palm-Vein Recognition

Kaveen Perera, Fouad Khelifi, Ammar Belatreche
Computer & Information Sciences Department
University of Northumbria at Newcastle, United Kingdom

## Abstract

Palm-vein recognition is a highly secure biometric modality due to the uniqueness and subcutaneous nature of vein patterns. However, low contrast in palm-vein images, caused by NIR light scattering and sensor limitations, remains a significant challenge. To address this, we propose the Intensity-Limited Adaptive Contrast Stretching with Bidirectional Gaussian-weighted Overlapping Tiles (ILACS-BGOT) method, an enhancement of the previously developed ILACS with Layered Gaussian-weighted Overlapping Tiles (ILACS-LGOT) technique. ILACS enhances local contrast, while BGOT mitigates blocky artefacts. This study further integrates RootSIFT features with KNN+RT and incorporates the previously introduced Mean and Median Distance (MMD) filter to investigate the parameter variations of both MMD and RT, and their impact on recognition performance. A comprehensive analysis was conducted across three benchmark datasets (CASIA, PolyU, and PUT), using 42 combinations of MMD filter thresholds and RT values. Results were evaluated using EER and Accuracy. Findings reveal that higher template sizes improve performance, while varying MMD thresholds reflect dataset-specific rotational variations. The proposed system demonstrates superior generalisability, achieving significant improvements in both EER and Accuracy over existing methods. Furthermore, the underlying ILACS-BGOT mechanism suggests potential applicability beyond palm vein recognition to other biometric modalities such as finger vein and palmprint recognition, and more generally to low-contrast image enhancement across computer vision applications.

## 1. Introduction

Palm vein recognition has emerged as a critical biometric modality, offering significant advantages for security and forensic applications. Unlike surface-based traits such as fingerprints or facial features, palm vein patterns reside beneath the skin, rendering them nearly impossible to forge or replicate [1], [2]. This intrinsic security, combined with the permanence of vein patterns throughout an individual's lifetime and their resilience to superficial skin conditions or injuries, positions palm vein recognition as a reliable and robust biometric solution. Moreover, its contactless and non-invasive nature addresses hygiene concerns while enhancing user convenience [3].

Despite these advantages, palm-vein images often suffer from low contrast due to both intrinsic and extrinsic factors related to the imaging process. The subcutaneous nature of palm veins limits the penetration of near-infrared (NIR) light, which is required to capture vein patterns, resulting in poor contrast between the veins and surrounding tissue [4]. Additionally, biological variations such as skin thickness, pigmentation, and moisture content, along with sensor limitations, contribute to inconsistent image quality across individuals and even across different regions of the same palm [5].

To illustrate this challenge, **Figure 1** [a] presents a low-contrast palm-vein image from the CASIA dataset [6], along with its corresponding histogram. Histograms provide a global representation of luminance distribution within an image, independent of the orientation or spatial arrangement of visual features [7]. As shown in **Figure 1** [b], the histogram demonstrates a narrow concentration of luminance values centred around the mid-tonal range, an indicator typical of low-contrast images [8]. This distribution confirms the visual impression of diminished contrast and highlights the need for effective contrast enhancement.

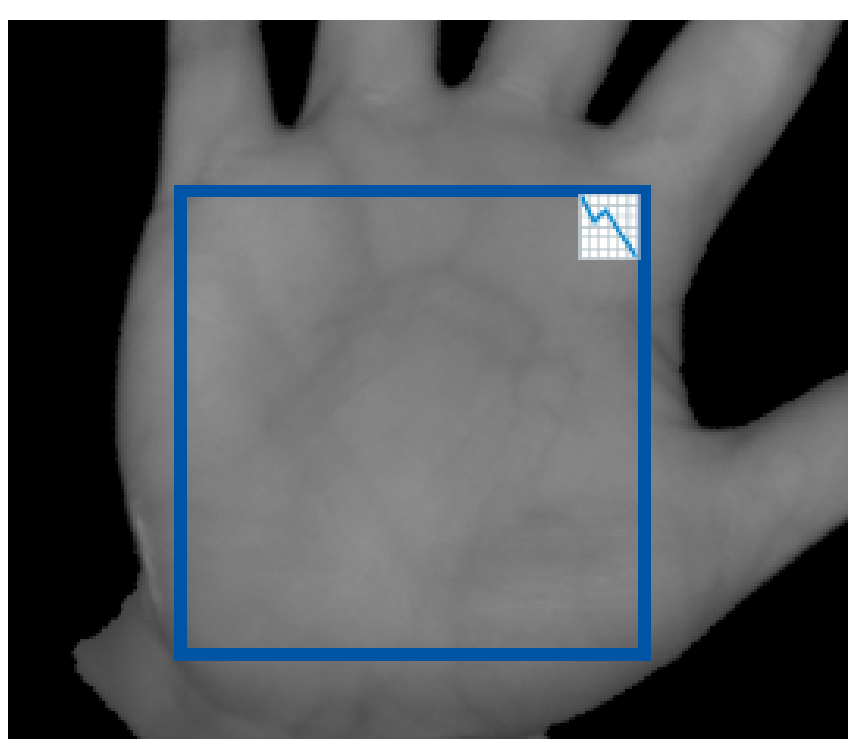

[a]: Sample image from the CASIA dataset

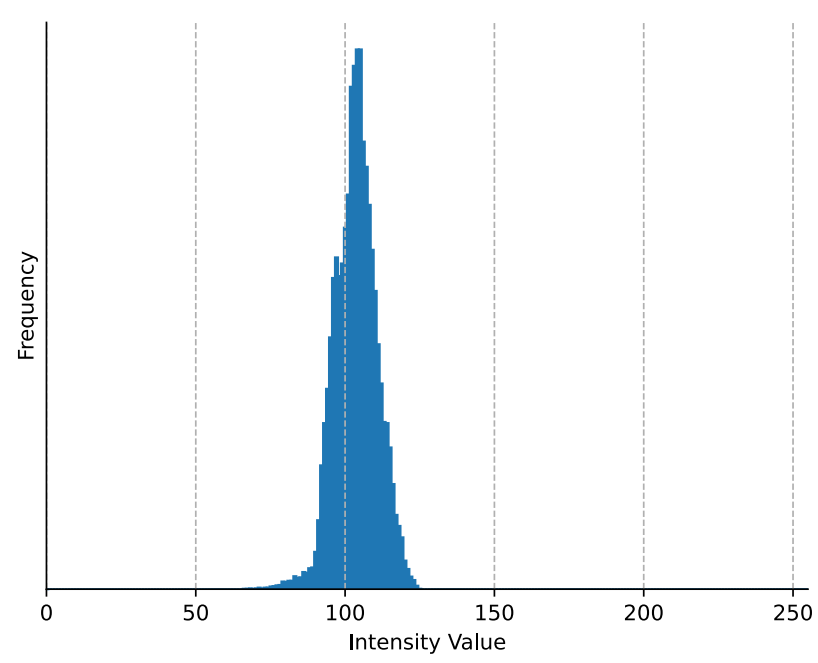

[b]: Histogram of [a]

**Figure 1**: Sample image from the CASIA dataset and the corresponding histogram.
ROI used to generate the histogram is marked with a blue rectangle.

Given these imaging challenges, a robust contrast enhancement (CE) method is essential for improving the visibility of vein patterns and enabling reliable feature extraction in palm vein recognition systems.

To address these challenges, the Multiple Overlapping Tiles (MOT) method, a novel contrast enhancement technique, was introduced in our previous study [9]. This method builds upon the concept of localised contrast enhancement, akin to Adaptive Histogram Equalisation (AHE) [10], while incorporating a limitation mechanism inspired by Contrast-Limited Adaptive Histogram Equalisation (CLAHE) [11].

The core concept of the MOT method is based on Intensity-Limited Adaptive Contrast Stretch (ILACS), a localized contrast stretch operation. MOT effectively addresses the blocky effect produced by applying the contrast stretch operation adaptively and was later renamed as ILACS Layered Gaussian-weighted Overlapping Tiles (ILACS-LGOT) in an extended preprint version of [9], made available as [12], to clearly attribute the operations involved. This extended version presents a more comprehensive description with visual representations, extended evaluations, and comparative examples that further support the performance of ILACS-LGOT relative to existing techniques.

The performance of ILACS-LGOT was evaluated using Equal Error Rates (EERs) with Euclidean Distance (ED)-based Scale-Invariant Feature Transform (SIFT) [13], [14] and RootSIFT [15] feature descriptor matching, employing filtering techniques such as KNN with Ratio Test (RT) [14], Bidirectional Matching, and RANSAC.

In [16], the Mean and Median Distance (MMD) filter was proposed to address the high number of false matches in SIFT-based feature matching. MMD applied geometric constraints on keypoints and was integrated into the ILACS-LGOT+RootSIFT+KNN+RT pipeline, significantly reducing false matches and further improving EER results.

However, limitations persisted. Although ILACS-LGOT's Gaussian-weighted blending was designed to smooth intensity discontinuities between adjacent tiles, this blending proved insufficient to fully eliminate the blocky effect. The residual artefact became apparent at larger tile sizes, where boundary mismatches between independently stretched regions are more pronounced; closer inspection confirmed the same effect present, to a lesser extent, at smaller tile sizes. This constrained the method's scalability and reduced its effectiveness across varying image resolutions and datasets.

To address these limitations, this study introduces the ILACS Bidirectional Gaussian-weighted Overlapping Tiles (ILACS-BGOT) method, designed to improve scalability for larger tile sizes and enhance adaptability across diverse datasets.

Building on previous work [9], [16], the proposed system is evaluated using three benchmark datasets: CASIA, PolyU, and PUT, which represent varying acquisition conditions. RootSIFT features, combined with KNN+RT+MMD filtering, are adopted based on their strong performance in previous experiments.

System performance is evaluated using two key metrics: EER and Accuracy. A comprehensive investigation of MMD thresholds and RT values is conducted to identify optimal parameters and examine generalisability across datasets.

The main contributions of this study are as follows:

- Proposes ILACS-BGOT, an enhancement of ILACS-LGOT (MOT), that improves local contrast and mitigates blocky artefacts, particularly at larger tile sizes.
- Comprehensive performance evaluation of the proposed method for palm-vein recognition using EER and Accuracy across three benchmark datasets (CASIA, PolyU, PUT).
- Investigates MMD thresholds and RT values across multiple datasets to provide generalisation guidelines for the proposed method.
- Compares results against existing palm-vein recognition systems in the literature to demonstrate improved robustness, accuracy, and generalisability.

The remainder of this article is structured as follows: Section 2 introduces key terminology and taxonomy. Section 3 reviews existing contrast enhancement techniques and related challenges. Section 4 details the proposed ILACS-BGOT method and provides a comparison against the ILACS-LGOT method. Note that ILACS-LGOT and ILACS-BGOT are evaluated under different experimental protocols (ROI extraction method, dataset composition), and direct EER comparison between the two is therefore not presented; the focus of this study is BGOT's performance relative to established palm-vein recognition systems. Section 5 describes the experimental setup. Section 6 presents and analyses the results. Section 7 concludes the findings and includes a public link to the ILACS-BGOT implementation.

## 2. Terminology and Taxonomy

To ensure consistency and clarity in the following sections, this part introduces key terms and a taxonomy that organises core concepts used throughout the study: specifically, gallery, probe, gallery-probe combinations, and template size.

A **gallery** refers to the set of stored biometric templates within a biometric system. Each gallery template represents an enrolled individual and consists of one or more biometric samples (e.g., images, fingerprints).

A **probe image** is a biometric sample captured during identification or verification and is used for comparison against the gallery. Unlike gallery images, probe images are not pre-enrolled but are acquired in real time. In a closed-set identification scenario, each probe is expected to correspond to a subject already represented in the gallery.

A **gallery-probe combination** defines a specific partitioning of biometric data for evaluation. In this setup, a subset of each user’s images is assigned to the gallery, while the remaining images are used as probes. By systematically rotating the roles of the images (e.g., using different subsets as gallery or probe), multiple combinations can be created, enabling robust cross-validation and performance analysis.

**Template size** refers to the number of images used to represent each subject in the gallery. A larger template size means more samples per individual, potentially improving matching accuracy but also affecting the size of the remaining probe set in a closed-set evaluation.

## 3. Related work

Wang and Wang [5] evaluated the impact of various CE methods on SIFT-based vein recognition, categorising the techniques into three groups: Linear/Nonlinear Functional Enhancements (LNFE), Histogram Equalisation and Its Variants (HEs), and Hierarchical Enhancement Models (HEM) or Multilayer Enhancement (MLE-HE).

LNFE methods, such as intensity normalisation (IN) and gamma correction (GC), improve visual quality but do not consider the neighbouring information. Hence, they often fail to adequately enhance the finer details required for accurate biometric matching. HEs techniques, including Adaptive-HE (AHE) and Contrast Limited-AHE (CLAHE), distribute intensity values more uniformly but are reported to perform poorly with the SIFT feature detector. Although adjusting CLAHE parameters can enhance the detection of SIFT keypoints, these adjustments can lead to amplified noise and visible blocky effects between the contextual regions.

HEM, which combines multiple techniques such as Difference of Gaussian (DoG) filtering and histogram equalisation (HE), offers improved contrast and detail preservation and performs well with SIFT feature detection. However, these methods can still discard vital subspace information necessary for robust feature extraction.

The most commonly used contrast enhancement technique in palm-vein recognition is CLAHE [17], [18], [19], [20], [21], [22], [23], [24], [25], [26], while other research has employed various multilayer HEM techniques to enhance the contrast of vein images. Zhou and Kumar [27] estimated the background intensity profile by subdividing the image into 32×32 size image tiles. To address the blocky effect, these image tiles overlap each other by three pixels. The average grey level of each block is then computed to estimate the respective background profile, which is scaled to match the original image using bicubic interpolation and subtracted from the original image. HE is then applied to enhance the vein image. Lee [28] employed a similar technique, subdividing the image into 16×16 non-overlapping tiles prior to calculating average greyscale values for each tile. However, they do not use HE as in [27]. Soh et al. [29] proposed a contrast enhancement process by applying a CLAHE operation and then stretching histograms of 8×8 image blocks. A common issue with these techniques is the blocky effect rendered at the edges of the image tiles. [29] and [28] do not address the blocky effect at all, while the 3-pixel overlapping used in [27] is not sufficient to address the blocky effect. Furthermore, using HE in [27] can clip out image details.

Yang and Zhang [30] worked upon the concepts of image dehazing techniques (which are used to remove scattered light effects on foggy images to improve background visibility) and introduced a new technique to reduce light scatter caused by skin tissues as a vein contrast enhancement method. Yang and Yang [31] proposed a multi-channel Gabor filter-based vein pattern enhancement method. Gabor filter parameters are automatically determined based on the width and orientation of vein networks; hence, the 4-orientations even-symmetric Gabor filter channels have different centre frequencies. Then, the four images are fused to produce the final enhanced image. Kim [32] proposes an image smoothing technique by accounting for the intensity change speed or rate. The rate of intensity change in flat areas where no vein patterns occur is slower than in the regions with vein patterns. The intensity difference between the original image and a smoothed version, which is referred to as smoothing speed, has been used to differentiate vein patterns from the background. A common issue observed with [30], [31], [32] is that these image smoothing techniques discard vital subspace information, which can be used to detect robust features.

Trabelsi et al. [33] calculated the global mean grey and variance values, then used a convolutional kernel which takes the global mean and variance along with an enhancement constant into account to calculate the local response. This method introduces a darkening effect in areas where vein patterns branch out or occur very closely to each other and amplifies the noise level, which would require further filtering.

Van et al. [34] proposed an LBP-based contrast enhancement method. They improved Center-Symmetric LBP (CS-LBP) to propose a new method referred to as Enhanced CS-LBP (ECS-LBP). They compared their contrast enhancement method against Gabor filters, LBP, and CS-LBP. These LBP-based contrast enhancement methods clip out and discard most of the subspace information, which makes them unsuitable to use with invariant features. Yan et al. [35] discussed an image sharpening technique in which they first multiplied the image with an enhancement factor of 10 and filtered the resulting image with the Prewitt high pass operator. This is then subtracted from the original image. Kang et al. [4] introduced a Difference of Gaussian filters (DoG) and histogram equalisation-based method. A DoG filter with a 4:1 Gaussian kernels ratio has been used as a band-pass filter to discard high frequencies. Then HE is applied to the DoG filtered image. This method is referred to as DoG-HE, which is also utilised by Yan et al. [36]. Similar to the issues discussed with other methods, these techniques discard subspace information, making them unsuitable for feature extraction.

Other studies have explored various methods for contrast enhancement in palm-vein recognition. Ahmed et al. [37] used Homomorphic filtering and Canny edge detection, while Ladoux et al. [38] combined noise reduction, brightness correction, and contrast normalisation, and Li et al. [39] used Gaussian and Median filtering for noise reduction. Parihar and Jain [40] used histogram equalisation and anisotropic diffusion filtering, and Soh et al. [29] used a low-pass Gaussian filter and Otsu thresholding. Van et al. [34] used Enhanced Center-Symmetric Local Binary Pattern (ECS-LBP), and Wu et al. [41] used a directional filter bank and Minimum Directional Code (MDC), and Yan et al. [35] employed sharpening filters.

# 4. The Proposed Method

The proposed Intensity-Limited Adaptive Contrast Stretching with Bidirectional Gaussian-weighted Overlapping Tiles (ILACS-BGOT) method can be classed into the Hierarchical Enhancement Model (HEM) [5], which employs localised contrast stretching to enhance local contrast more effectively and a hierarchical approach to eliminate the blocky effect.

The core concept of the proposed method is based on ILACS, a localized contrast stretch operation. In our previous work [9], we introduced the basic formulation of this method. BGOT and LGOT are methods introduced to address the blocky effect produced by ILACS. Since both methods are inherently based on ILACS, we will use the terms BGOT and LGOT in subsequent sections when referring to their respective implementations.

In our previous work [9], [42], we introduced the formulation of ILACS. This section revisits that foundation, as the current method builds upon it and a review will aid the reader's understanding of the underlying concepts. The LGOT method is then briefly discussed, followed by the introduction of the BGOT algorithm, and a concluding comparison between the two.

## 4.1 Intensity Limited Adaptive Contrast Stretch

Adaptive Contrast Stretching (ACS) improves the contrast of palm-vein images by segmenting the image into $N \times N$ square tiles and stretching the contrast within each tile to maximise the available dynamic range, as defined by ***Equation (1)*** [43]. The resulting output is then rescaled to the 0–255 range by multiplying by 255 to match the 8-bit palm-vein images used in this study.

Following are the steps taken to generate an ACS image:

- Subdivide the image into square tiles.
- Find $\min(T_N)$, the minimum pixel values of the particular $T_N$ tile.
- Find $\max(T_N)$, the maximum pixel values of the particular $T_N$ tile.
- Apply the $F(T_{N_i})$ in ***Equation (1)*** for each pixel denoted as, $T_{Ni}$.
- Repeat all the above steps for all $T_N$ tiles.

However, this approach may be sensitive to noise and artefacts. Similarly, AHE encounters the same issue, as it tends to intensify noise in regions of an image with nearly uniform intensity [44]. To mitigate this, CLAHE incorporates a clipping mechanism that defines a threshold, restricting the maximum number of pixels allowed in any histogram bin within each tile [11]. A multiplier was proposed in [43] to scale down the contrast stretch, serving as a control measure to reduce noise and artefacts.

$$F(T_{N_i}) = \left(\frac{T_{N_i} - \min(T_N)}{\max(T_N) - \min(T_N)}\right) \times 255 \quad (1)$$

Where:

$N$ = length of a tile in pixels,
$T_N = N \times N$ sized image tile,
$i$ refers to an individual pixel in the image tile $T_N$.

To ensure a fully localised operation, the denominator of ***Equation (1)*** was integrated with the multiplier and applied using a floor function to scale down the contrast stretch. This functions as a stepping mechanism, automatically adjusting the scaling down process based on the local pixel intensities within each image tile. This method is referred to as Intensity Limited Adaptive Contrast Stretch (ILACS). The modified formula of the ILACS method is presented in ***Equation (2).***

$$F(T_{N_i}) = \left(T_{N_i} - \min(T_N)\right) \times \left\lfloor \frac{255}{\max(T_N) - \min(T_N)} \right\rfloor \quad (2)$$

The symbol $\lfloor\ \rfloor$. in ***Equation (2)*** indicates the floor function.
The floor function, $\lfloor x \rfloor$, is defined as the greatest integer $n$ such that $n \le x$:

$$\lfloor x \rfloor = max\{n \in \mathbb{Z} \mid n \le x\}$$

If the input image contains sections of uniform colours, such as a dark background, a division-by-zero error can occur due to the denominator in ***Equation (2)*** reaching zero. This issue is mitigated by employing a binary mask, which ensures that the contrast stretch operation is applied only to the areas where the palm image is present. Alternatively, if the range of pixel values within a tile evaluates to zero, the value of the denominator can be set to one.

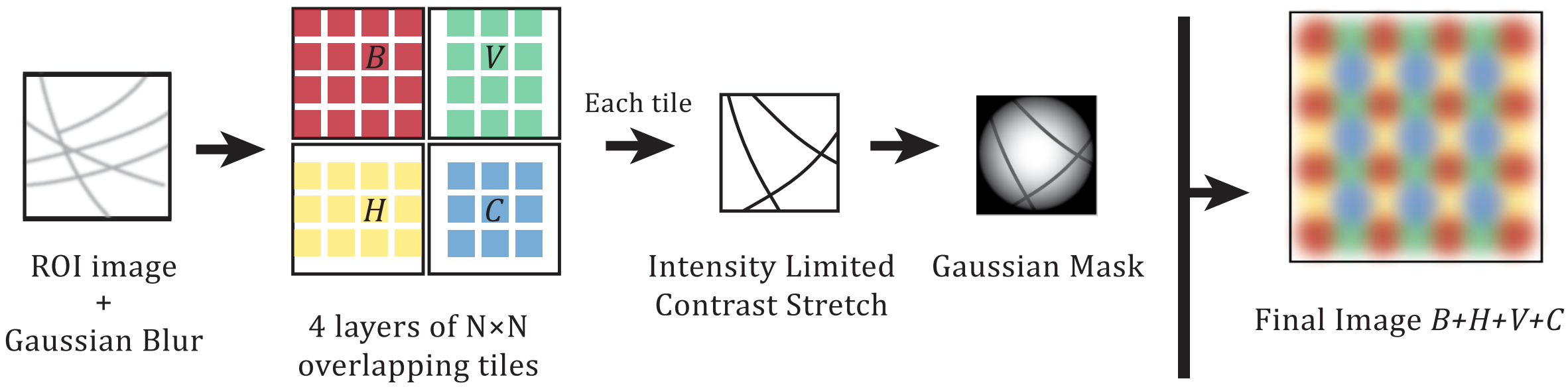

**Figure 2**: The flowgraph of the ILACS-LGOT method.
(Gaps between the tiles added for visual clarity and illustrative purposes only).

ILACS functions as a point operator, meaning it does not consider the intensities of neighbouring tiles. Consequently, two adjacent pixels with slight intensity differences but located in separate tiles may be remapped to significantly different intensity values, as the transformation depends solely on the minimum and maximum pixel values within each respective tile. This can result in a sharp edge or blocky effect at tile boundaries. This issue is a well-documented challenge in localised image processing, where methods must strike a balance between enhancing local contrast and maintaining global consistency.

### 4.2 Layered Gaussian-weighted Overlapping Tiles (LGOT)

AHE mitigates the blocky effect by employing methods that incorporates multiple sampling techniques to smooth transitions between tiles. In AHE, sampling can be performed at intervals equal to the contextual region size (mosaic sampling), at half the contextual region size (half-sampling), or at every pixel (full-sampling). Some variations include weighted AHE, which assigns greater weight to pixels closer to the centre of the contextual region [10].

Zhou and Kumar [27] utilised a technique involving 32×32 image tiles with a 3-pixel overlap. The average grey level for each tile was computed to estimate a background profile, which was then resized to match the original image dimensions using bicubic interpolation. This profile was subsequently subtracted from the original image, followed by the application of HE. However, a 3-pixel overlap is insufficient to effectively overcome the blocky effect.

The LGOT method builds upon the strengths of AHE and enhances the contrast stretch approach by introducing multiple overlapping layers with half-sampling. Three additional layers of image tiles are generated: one centred along the horizontal edges of the base layer tiles, another centred along the vertical edges, and a third centred at the vertices where the tiles intersect. **Figure 2** illustrates the flowgraph of the ILACS-LGOT method.

The final intensity value of each pixel is computed as a weighted sum of the intensity values from all overlapping tiles that include the pixel. These weights are determined using a Gaussian mask, which seamlessly blends the overlapping regions, ensuring a smooth transition between tile boundaries. This process enhances contrast while producing a visually consistent image enhancement.

Further details about the LGOT method are provided in [9] and [42].

Upon completing the initial research phase using the LGOT method on the CASIA dataset and comparing it with other methods, it was observed that it does not perform optimally with larger tile sizes due to the limitations of automatically setting the standard deviation of the Gaussian alpha mask. Several experiments were conducted with different variations of LGOT, leading to the development of a simplified approach. This new approach can render a contrast-stretched image with larger tile sizes without any blocky effects or edges.

### 4.3 Bidirectional Gaussian-weighted Overlapping Tiles (BGOT)

In contrast to the four-layers approach in LGOT, the new method employs an array of bidirectional overlapping tiles. **Figure 3** depicts the sequence of operations involved in processing the image tiles and blending their overlapping regions in the ILACS-BGOT method.

The formula for the two-dimensional (2D) isotropic Gaussian function used to generate Gaussian-weights for pixels in each image tile in both LGOT and BGOT methods is given in ***Equation (3)*** [45]. In ***Equation (3)***, $x$ and $y$ are the width and height of an image tile, and $\sigma$ is the standard deviation.

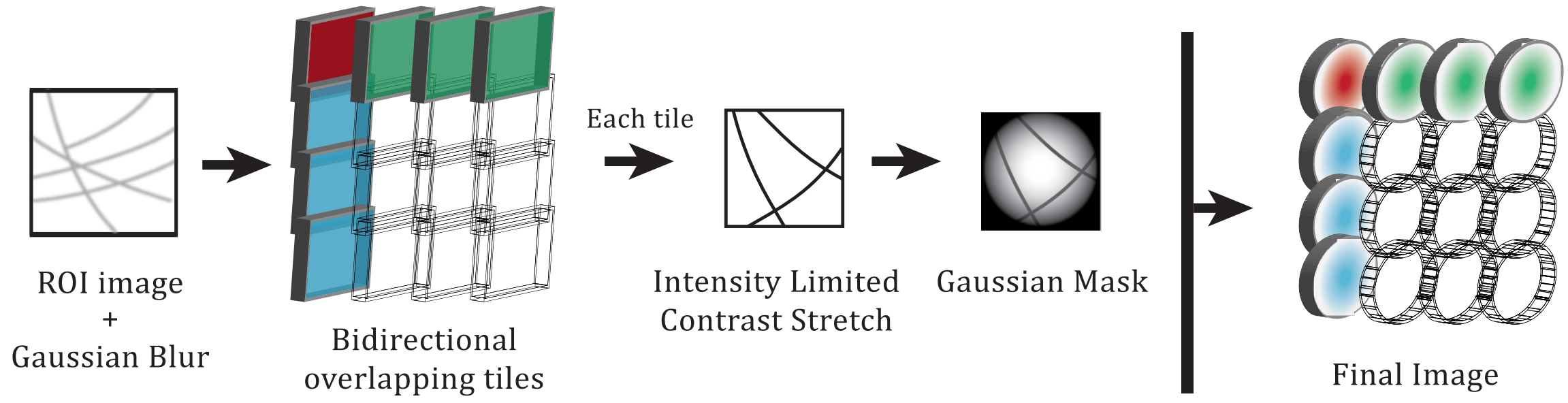

**Figure 3**: The flowgraph of the ILACS-BGOT method.
(Gaps between the tiles added for visual clarity and illustrative purposes only).

$$G(x, y) = \frac{1}{2\pi\sigma^2} e^{-\frac{x^2+y^2}{2\sigma^2}} \quad (3)$$

These weights act as a transparency or alpha channel ($\alpha$) for each layer [46].Then ***Equation (4)*** is used to blend each tile $f$ using a greyscale alpha channel α over a background image $b$ to create a new image $I_N$. This is also referred to as the over-operator [47]. This formula assumes the background to be a solid image without an applied alpha channel.

$$I_N = [(1 - \alpha)\, b] + \alpha\, f \quad (4)$$

To explain the implementation of the BGOT algorithm, a column-first (or column-major) order is adopted. **Figure 4** depicts the tile overlaying of the BGOT method presented in column-first order. Alternatively, a row-first (or row-major) order may be used, producing equivalent outcomes.

The process begins with a blank background (a 2-dimensional array with all values set to 0) with the same dimensions as the original image. The first tile (*'A'* on **Figure 4**) is computed with ILACS and added or blended onto the background using ***Equation (4)*** utilising the radial Gaussian mask generated with ***Equation (3)*** as the alpha channel.

The BGOT methods employ square tiles, and the length of the tile is denoted as $N$ in ***Equation (2)***. It was observed that the standard deviation or falloff of $\sigma = N/5$ for the Gaussian mask produces consistent results across small to large image tile sizes, tested up to 256×256 pixels. This value can be adjusted to accommodate even larger tile sizes if required. This procedure is then repeated on vertically overlapping tiles until the last tile (*'B'* on **Figure 4**) along the left edge of the image is reached, resulting in the first column of alpha-blended tiles.

The second column of tiles (starting from tile *'C'* on **Figure 4**) is horizontally displaced to overlap with the first column. Each column of tiles overlaps with the next column of tiles. Similar to the first column of tiles, the alpha blending operation is repeated on every tile until the final tile of the image (tile *'D'* on **Figure 4**) is reached.

The displacement of the overlapping tiles is computed by dividing the tile length $N$ by the number of overlapping tiles (denoted as $r$) that occupy the space of the initial tile (*'A'* on **Figure 4**) along the first column. Consequently, each overlapping tile is displaced by $N/r$ pixels along a specified axis. In the example illustrated in **Figure 4**, $r$ equals 2. The maximum value that $r$ can reach is the tile length $N$, while the maximum number of overlapping tiles that a single pixel could present is $r^2$. Optimal results were observed when $r$ was set to 4.

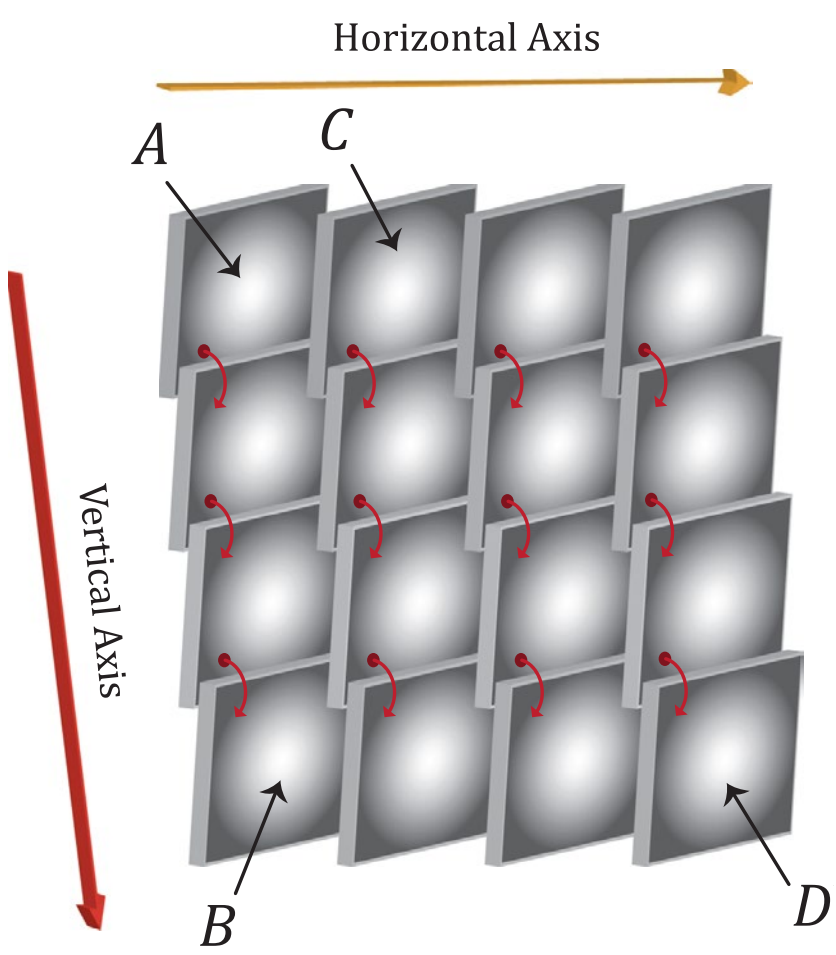

**Figure 4**: Illustration of the BGOT algorithm in a column-first order.
The process starts with a blank background of the same dimensions as the original image, then each tile in the first column of the vertical axis starting from tile 'A' until 'B' is computed with ILACS and blended to the background employing a radial Gaussian mask. This procedure is repeated on all overlapping tile columns horizontally, until reaching tile 'D' covering the entire image. Similar results can be achieved using a row-first order.

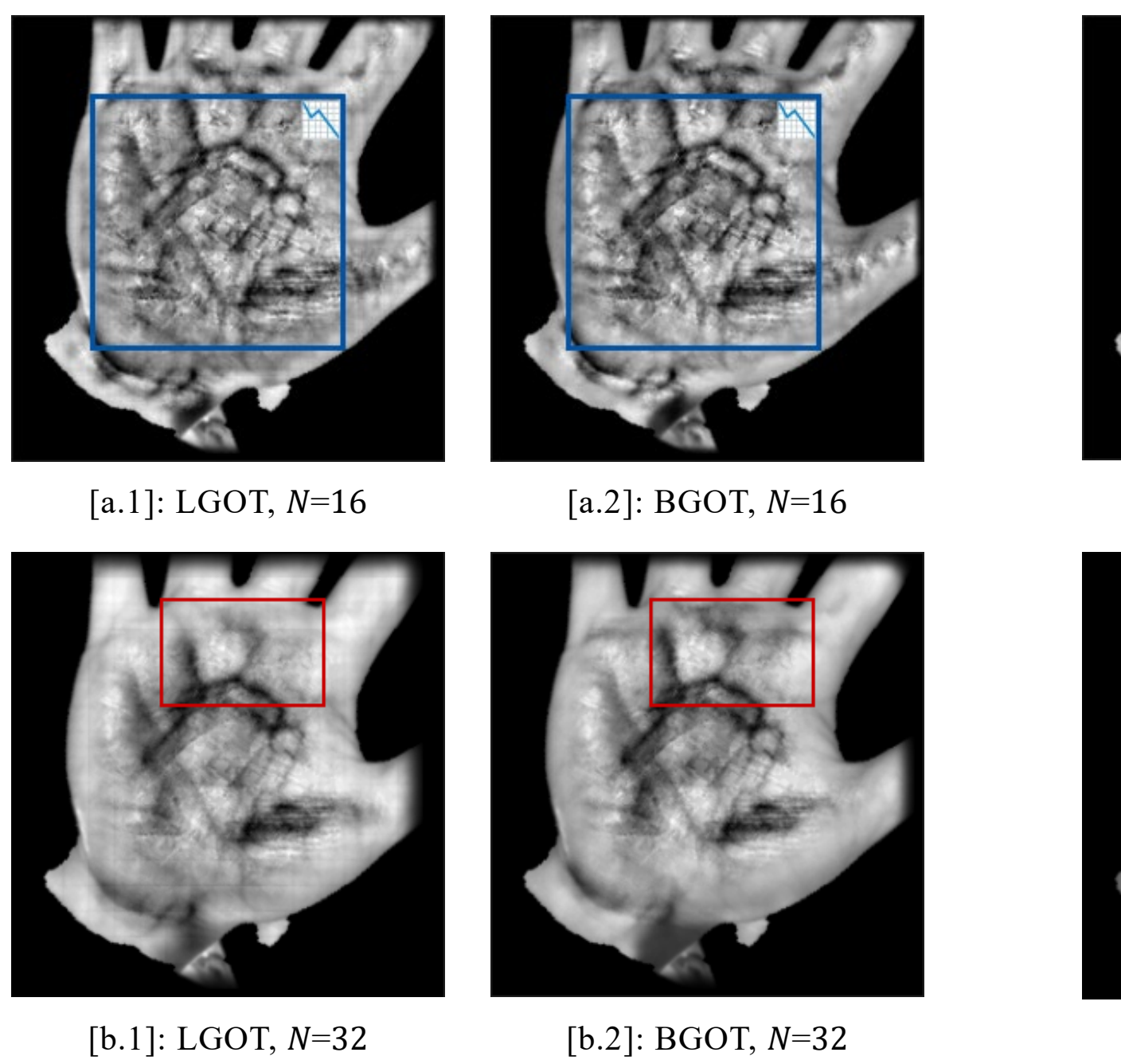

**Figure 5**: Comparison of LGOT and BGOT methods with tile sizes 16 and 32 pixels.

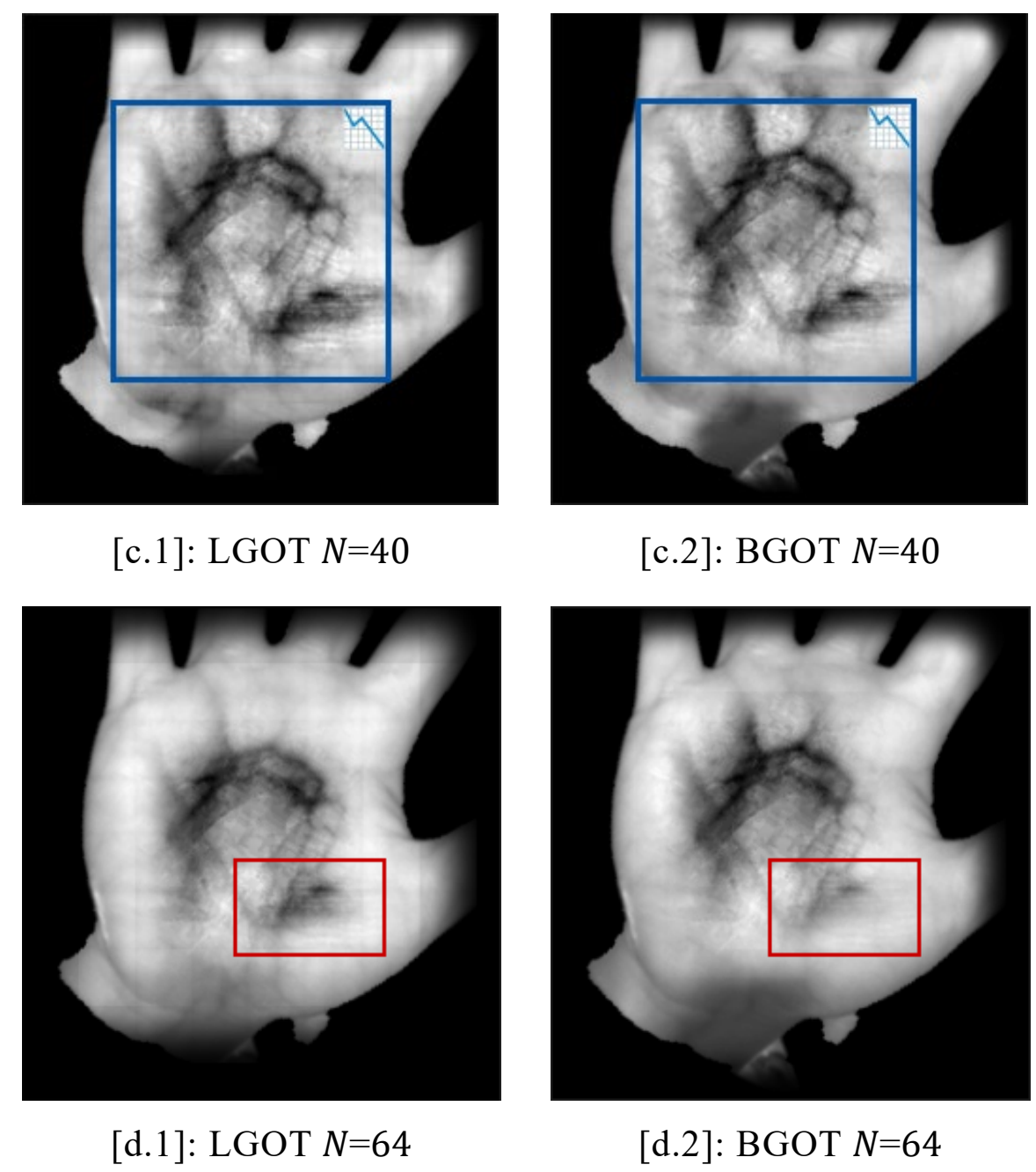

**Figure 6**: Comparison of LGOT and BGOT methods with tile sizes 40 and 64 pixels.

## 4.4 Comparison: LGOT vs. BGOT

**Figure 5** and **Figure 6** present four comparisons between the LGOT and BGOT methods utilising different tile sizes. A common observation across these comparisons is that the pixels surrounding the vein patterns appear slightly darker in the BGOT images.

This darkening effect can be observed by comparing the areas inside the red rectangles of **Figure 5** [b.1] and [b.2]**.** However, in certain areas (as can be seen close to the bottom edges of the indicated ROI squares on **Figure 5** [a.1] and [a.2]), the BGOT method has further improved the details and made the skin surface appear brighter than the LGOT method.

**Figure 5** [a.1] and [a.2] compare the LGOT and BGOT methods when the tile size is set to 16, while **Figure 7** [a.1] and [a.2] present their histograms on the indicated ROI.

In the histogram in **Figure 7** [a.1], the frequency peak occurs when the intensities are proximate to 150, while in **Figure 7** [a.2], it is near 175. The BGOT method has increased the representation in the range of medium intensities, slightly reducing the representation in the brighter intensities range.

**Figure 6** [c.1] and [c.2] compare the LGOT and BGOT methods when the tile size is set to 40. **Figure 7** [c.1] and [c.2] present their respective histograms on the indicated ROI. When comparing these histograms, it can be observed that the frequency peak has shifted in the opposite direction for the BGOT method compared to the previous example, while the representation in the high-intensity range has been reduced.

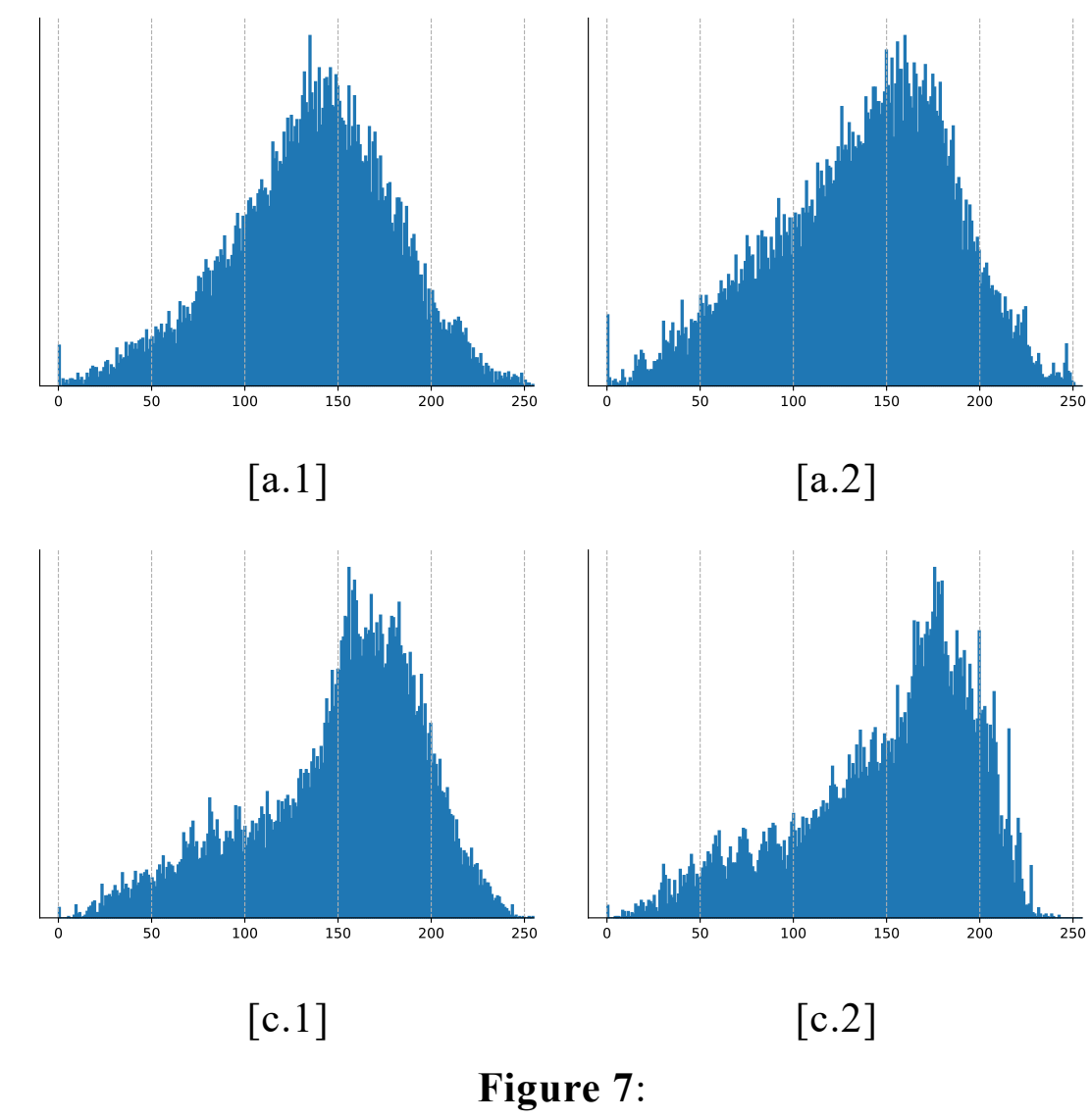

**Figure 7**:
[a.1] and [a.2] - Corresponding histograms for subfigures [a.1] and [a.2] of Figure 5.
[c.1] and [c.2] - Corresponding histograms for subfigures [c.1] and [c.2] of Figure 6.

Furthermore, it can be observed by comparing the areas inside the red rectangles of **Figure 6** [d.1] and [d.2], that the self-shadow areas have been improved with the BGOT method.

The LGOT method can produce a shallow edge pattern when using larger tile sizes above 32 pixels. This effect can be observed in **Figure 5** [b.1], **Figure 6** [c.1] and is highly noticeable in **Figure 6** [d.1]. In contrast, the BGOT method does not produce such artefacts and yields consistent results across various tile sizes.

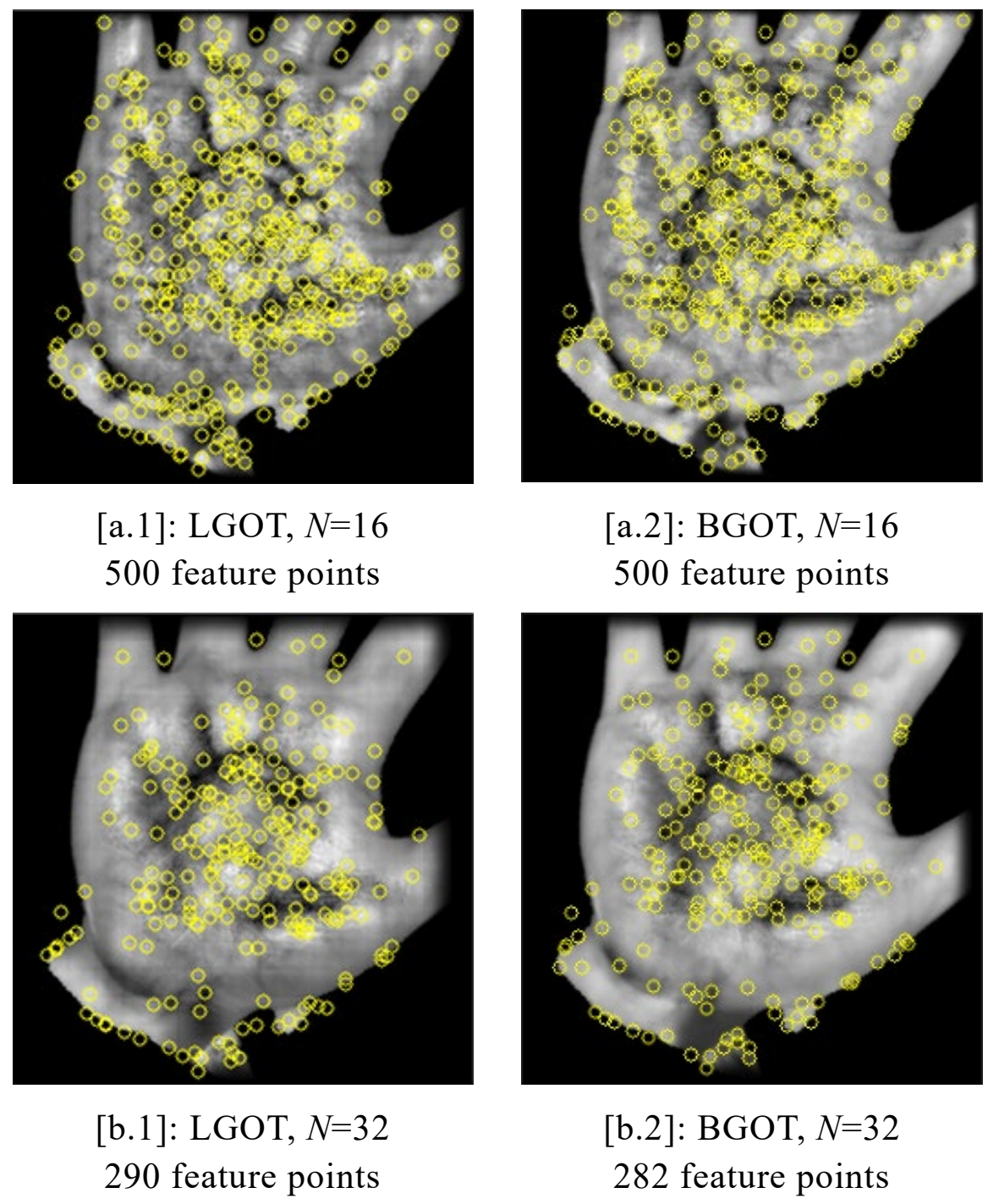

[a.1]: LGOT, $N$=16
500 feature points

[a.2]: BGOT, $N$=16
500 feature points

[b.1]: LGOT, $N$=32
290 feature points

[b.2]: BGOT, $N$=32
282 feature points

**Figure 8**: SIFT features detection comparison between LGOT and BGOT methods with tile sizes 16 and 32 pixels.

**Figure 8** presents comparisons of SIFT feature detection between the LGOT and BGOT methods. When a smaller tile size of 16 is employed (as shown in **Figure 8** [a.1] and [a.2]), both methods detect the maximum set limit of 500 SIFT keypoints. However, when larger tile sizes are used, the BGOT method produces slightly fewer keypoints than the LGOT method. Upon closer inspection, it was revealed that some keypoints were detected on shallow edge pattern artefacts in the LGOT method. These keypoints can be discarded during the filtering process.

## 5. Evaluation Framework

### 5.1 Experimental Setup

In [9], the performance of the LGOT method was assessed using the CASIA dataset, employing a full palmar ROI, pre-aligned to minimise rotational and translational variations. LGOT enhanced the contrast of palm vein images, improving SIFT feature extraction. Results demonstrated that LGOT effectively reduced artefacts such as blocky effects while preserving vein patterns, establishing it as a robust enhancement method for palm vein recognition.

Performance was measured by comparing EERs obtained with feature matching and various filtering techniques such as KNN and RT. In [16], MMD was evaluated as an additional false matches filtering step within the LGOT enhanced pipeline on the CASIA dataset, using SIFT and RootSIFT descriptors. Results showed that MMD significantly reduced false matches by incorporating geometric constraints on keypoints, improving recognition performance.

Building upon these foundations, this section evaluates the enhanced BGOT method and extends the application of MMD to provide comprehensive testing across multiple metrics and datasets. The evaluation focuses on determining the optimal parameters and assessing generalisability across datasets.

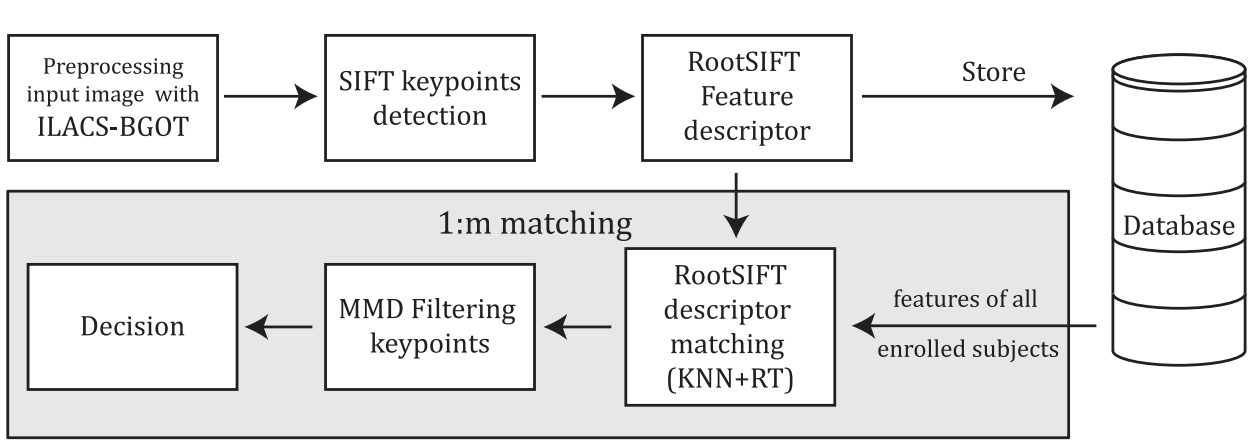

**Figure 9**: Flowgraph of the proposed palm-vein recognition system using the MMD filter.

The flowgraph of the palm-vein recognition system used for evaluation which incorporates the BGOT as the image enhancement method and the MMD filter is presented in **Figure 9**. 16×16 pixels image tiles are employed with the BGOT method to maintain consistency with previous studies [9], [16], and the detected SIFT feature descriptors are converted into RootSIFT descriptors.

RootSIFT is a modification of the standard SIFT descriptor designed to improve matching accuracy and robustness. Proposed by Arandjelović and Zisserman [15], RootSIFT transforms the SIFT descriptor by applying a square root operation to each element after normalisation. This transformation addresses the skewness in the original SIFT descriptor, where larger gradient values could dominate and overshadow other features. The square root operation effectively converts the Euclidean distance between RootSIFT descriptors into an approximation of the Hellinger distance, which is more suitable for comparing histograms. This modification enhances the robustness of feature matching under varying image conditions, without adding a significant computational overhead.

### 5.2 Ratio Test (RT)

To refine the matching process and reduce false positives, the distance ratio test (RT) introduced by Lowe [14] adds a layer of precision to the matching strategy. Instead of simply identifying the closest match by the smallest ED, the RT involves calculating the ratio of the distance to the nearest neighbour and the distance to the second-nearest neighbour. A keypoint match is considered valid if the ratio of the smallest to the second-smallest ED is below a certain threshold, commonly set between 0.7–0.8 as suggested by Lowe. This 'Ratio Test' ensures that the selected match is distinctly closer than the next best match, thereby increasing the confidence in the robustness of the match. This method significantly reduces the likelihood of false matches by ensuring that the matches are not only close but also uniquely closer than any other potential matches. In this setup, RT was applied to 2 nearest neighbour features matched using the KNN algorithm.

### 5.3 The MMD algorithm

The MMD filter is based on the geometric placement of SIFT keypoints, measuring the pixel distance between the $\mathcal{X}$ and $\mathcal{Y}$ coordinates of matching keypoint pairs. It operates on the hypothesis that false positive matches should have greater distances than true positive matches, and that the $\mathcal{X}$ and $\mathcal{Y}$ coordinate distances of true matches should be minimal.

In an ideal scenario, when palm-vein images are perfectly pre-aligned on a Cartesian plane, true positive matches should have identical $\mathcal{X}$ and $\mathcal{Y}$ coordinates. However, in practice, variations such as hand positioning, image noise, rotation, and scale differences prevent perfect alignment, resulting in nonzero distances even for true matches. The MMD filter is designed to accommodate these rotational and scale variations present between image pairs, ensuring more reliable matching.

This distance is not the same as the Euclidean Distance (ED) used in SIFT feature matching. ED, used by methods such as KNN and RT, measures similarity between 128-dimensional SIFT feature descriptors based on the sum of squared differences between vector elements. The MMD filter, however, focuses on the spatial arrangement of keypoints rather than descriptor similarity.

When matching SIFT features between two palm-vein images ($\mathcal{G}$ and $\mathcal{P}$), the horizontal ($x$) and vertical ($y$) distances ($\mathcal{D}_{ix}$, $\mathcal{D}_{iy}$) between each matched keypoint pair ($\mathcal{M}_i$) are measured. The mean ($\mu_x$, $\mu_y$) and median ($\tilde{x}_x$, $\tilde{x}_y$) distances are then calculated for both axes. Next, the number of matching pairs $\mathcal{N}_L$ (when $\mathcal{D}_{ix} \leq \mu_x$ and $\mathcal{D}_{iy} \leq \mu_Y$), $\mathcal{N}_H$ and (when $\mathcal{D}_{ix} > \mu_x$ and $\mathcal{D}_{iy} > \mu_y$) is counted based on their distances. To account for rotation and scale variations, two thresholds are introduced: maximum mean threshold ($T_\mu$) and maximum distance threshold ($T_\mathcal{D}$).

A pair of palm-vein images is considered a match if at least one of the following conditions is met:

- If $\mathcal{N}_L \geq \mathcal{N}_H$
- If $\mu_x \leq T_\mu$ **and** $\mu_y \leq T_\mu$
- If $\tilde{x}_x \leq \mu_x$ **and** $\tilde{x}_y \leq \mu_y$

If the image satisfies the selection criteria outlined above, a pair of keypoints is classified as a true positive match only if both distances are lower than their respective means and the maximum distance threshold for both the horizontal ($\mathcal{D}_{ix} < \mu_x$ **and** $\mathcal{D}_{ix} < T_\mathcal{D}$) and vertical ($\mathcal{D}_{iy} < \mu_Y$ **and** $\mathcal{D}_{iy} < T_\mathcal{D}$) axes.

The algorithm's notations and comparisons against existing research can be found in [16]. An extended preprint version of this study is made available as [48], presents additional explanations, including detailed examination along with extra sample cases.

### 5.4 Datasets and ROI

Three benchmark datasets: PolyU [49], PUT [50] and CASIA [6], representing varying acquisition conditions and dataset characteristics are utilised in this study. The full datasets were used in a one-to-many (1:*m*) closed-set approach.

The PolyU dataset contains images from 250 individuals, 500 different palms, employing multispectral imaging under blue (470nm), green (525nm), red (660nm), and NIR (880nm) illuminations. For this research, only the 6000 images from the NIR subset are used. The dataset was captured using a guided contact setup where participants had to align the valley point between the middle and ring finger against a guide, resulting in very minor placement-related variations between the palm images. For this study, the PolyU dataset was obtained as cropped ROI images, each measuring 128 × 128 pixels.

The PUT NIR palm-vein dataset comprises images of 100 palms from 50 volunteers, capturing 1,200 palm-vein images from both left and right hands in 880nm wavelength. The dataset was captured using an unguided contact setup, where subjects freely placed their palms over the acquisition window. Images were obtained as 1280 × 960 cropped bitmaps from the central area of palms and saved in BMP format. The images were resized to 340 × 256 pixels by fixing the height at 256 pixels and scaling the width proportionally before rounding it down to the nearest even pixel, producing a resolution that closely approximates the original 4:3 aspect ratio.

The CASIA multi-spectral dataset was captured under wavelengths of 460nm, 630nm, 700nm, 850nm, 940nm, and white light in 788 × 576 resolution. A total of 1,200 images are available for each wavelength in JPEG format, comprising six images per left and right hand, with varying orientations and stretch levels (of fingers and thumb) captured from 100 participants, resulting in 200 different palms.

To evaluate the generalisability of the proposed method, uniform test settings must be applied across all datasets. The PolyU and PUT datasets were obtained in a cropped format, whereas, in previous experiments [9], [16], the full palm area of the 850nm CASIA subset was utilised.

To maintain consistency across datasets used in this study, a cropped ROI for the 850nm subset of the CASIA dataset has been adopted. This cropped ROI is structured around a 2D×2D square (refer to **Figure 10** [a]), where 'D' is defined by the distance between the valley point $P_R$ of the index and middle fingers and the valley point $P_L$ between the ring and little fingers, also known as the pinky finger.

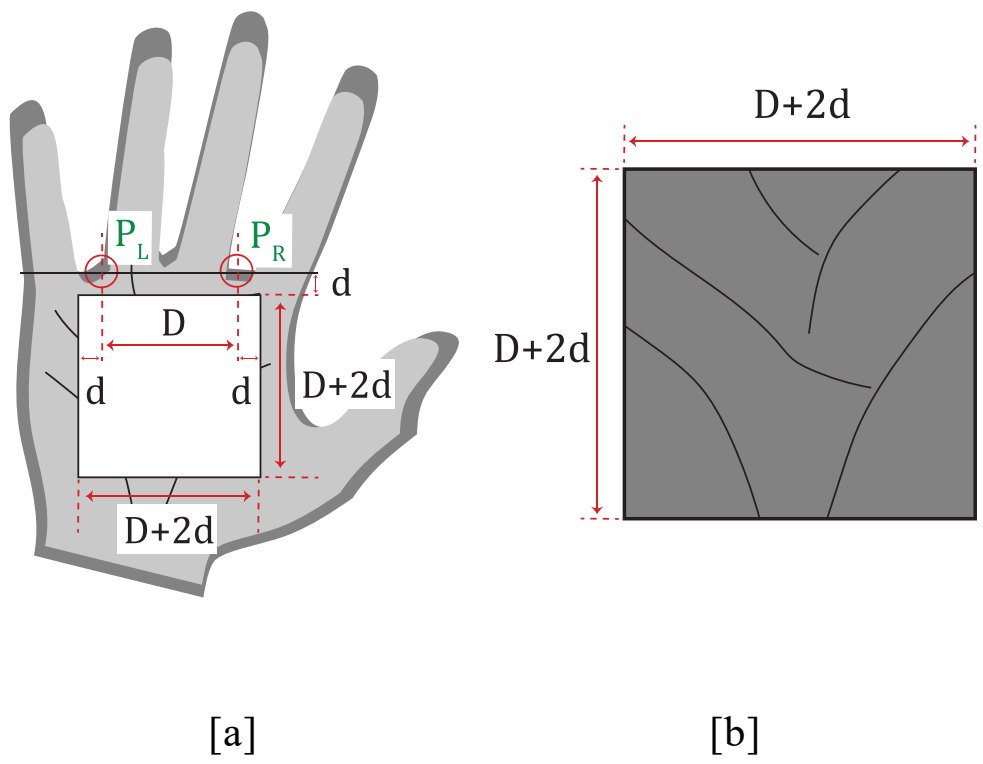

[a] [b]

**Figure 10**: A square ROI from the CASIA dataset is employed in evaluation Stage Two (d = D/6).

It was observed that the approximate width of a finger at its base is about D/3. Half the approximate width of a finger, denoted as 'd' (d = D/6), is used for offsetting and padding the ROI.

The ROI image is then cropped to the boundaries of the padded square, with each side measuring D+2d (**Figure 10** [b]), and subsequently resized to 128 × 128 pixels while maintaining a 1:1 aspect ratio.

### 5.5 Parameters Range and Gallery-Probe Combinations

The proposed BGOT method was tested using template sizes of 1 to 5 with KNN+RT+MMD filter combination. To comprehensively investigate MMD filter thresholds and their corresponding RT values, experiments were conducted on each dataset using six RT values (0.65, 0.7, 0.75, 0.8, 0.85, and 0.9) and seven MMD thresholds (5, 10, 15, 20, 30, 40, and 50). This resulted in a total of 42 mean EER and mean Accuracy values for each template size. These results were analysed to determine the optimal parameters and assess their generalisability across datasets. It should also be noted that when the RT value approaches 1 (specifically beyond 0.95), the matching process effectively reverts to ED-based matching.

Initial experiments indicated that different gallery-probe combinations yield varied results for the same subject, necessitating the testing of all combinations to evaluate and cross-validate the performance of the proposed system effectively. Although this extensive testing is computationally intensive, it is essential to properly understand how the parameters behave and assess the potential to generalise this method to practical applications and other datasets.

Therefore, all the results are produced as mean values calculated using all the gallery-probe combinations. These results provide a more robust evaluation of the biometric system's performance than using a single gallery-probe combination. All the decision rates are produced as mean values calculated using all the gallery-probe combinations. For multiple templates, the sum of matched features between each image in the template and the probe was used.

### 5.6 Number of Calculations

The experiments are computationally intensive due to the large number of comparisons and parameter variations. Each image is compared against every other image in the dataset, and multiple parameter combinations (six RT values and seven MMD thresholds) further increase the total number of operations.

The total number of calculations per ROI size is determined based on the dataset size and the combinations of RT and MMD threshold values. For the PolyU dataset, which contains 6,000 images, the total number of calculations per ROI size exceeds 1.5 billion (6000×5999×6×7=1,511,748,000). For the PUT and CASIA datasets, each containing 1,200 images, the total number of calculations per ROI size exceeds 60 million per dataset (1200×1199×6×7=60,429,600).

### 5.7 Evaluation Metrics Used

This study employs two primary performance evaluation metrics: EER, Accuracy, and utilises RootSIFT features to conduct a comprehensive investigation.

The Equal Error Rate (EER), frequently reported in biometrics research literature, provides a balanced measure of a system's performance by equating the false acceptance rate (FAR) and false rejection rate (FRR). EER is measured when FAR = FRR [51]. However, these metrics are applicable only when measuring the performance of a complete biometrics system, not just the matching algorithm. FRR is also referred to as a Type I error and FAR as a Type II error in various literature, which includes Failure to Acquire (FTA) and Failure to Enrol (FTE) rates [52]. Some literature, which focuses solely on matching algorithms, uses FAR synonymously with FPR and FRR with FNR. However, FAR and FRR are system-level errors that consider the total number of transactions. When evaluating only a matching algorithm, EER should be measured when the False Positive Rate (FPR) equates to the False Negative Rate (FNR). FPR and FNR are matching algorithm errors which do not consider FTA or FTE [52], [53], [54]. FPR and FNR are produced as mean values and mean EER is produced when the mean FPR equates to mean FNR.

Accuracy or the Identification Accuracy (ACC) is a commonly used performance measure in biometrics research. Accuracy of a matching algorithm is the ratio of correct classifications to the total number of samples [53].

## 6. Results and Analysis

### 6.1 EER and Accuracy Extremes Across RT and MMD Combinations

**Table 1**, **Table 3**, and **Table 5** present the lowest mean EER and the highest mean Accuracy values obtained out of the 42 RT and MMD parameter combinations for each template size, for CASIA, PolyU, and PUT datasets, respectively, alongside their corresponding RT values and MMD thresholds. **Table 2,**

**Table 4**, and **Table 6** present the lowest EER and highest Accuracy values recorded with any single gallery-probe combination for the respective datasets.

**Table 1**: Lowest **mean EER %** and the highest mean Accuracy % values on the CASIA dataset.

| Template Size | Mean EER % | RT Val. | MMD Tr. | Mean Accuracy % | RT Val. | MMD Tr. |
|---|---|---|---|---|---|---|
| 1 | 4.58 | 0.9 | 10 | 94.57 | 0.85 | 10 |
| 2 | 2.56 | 0.9 | 10 | 97.3 | 0.85 | 10 |
| 3 | 1.08 | 0.85 | 10 | 98.72 | 0.85 | 10 |
| 4 | 0.63 | 0.85 | 10 | 99.35 | 0.85 | 10 |
| 5 | 0.43 | 0.9 | 10 | 99.52 | 0.9 | 10 |

**Table 2**: Lowest EER % and highest Accuracy % values recorded from any combination from CASIA.

| Template Size | Lowest EER % | RT Val. | MMD Tr. | Highest ACC % | RT Val. | MMD Tr. |
|---|---|---|---|---|---|---|
| 1 | 3.51 | 0.85 | 10 | 95.78 | 0.85 | 10 |
| 2 | 1.12 | 0.85 | 15 | 98.71 | 0.85 | 15 |
| 3 | 0.36 | 0.8 | 50 | 99.63 | 0.9 | 10 |
| 4 | 0 | 0.85 | 10 | 99.86 | 0.85 | 10 |
| 5 | 0 | 0.8 | 5 | 100 | 0.8 | 5 |

**Table 3**: Lowest **mean EER %** and the highest mean Accuracy % values on the PolyU dataset.

| Template Size | Mean EER % | RT Val. | MMD Tr. | Mean Accuracy % | RT Val. | MMD Tr. |
|---|---|---|---|---|---|---|
| 1 | 3.24 | 0.85 | 5 | 95.83 | 0.85 | 10 |
| 2 | 1.74 | 0.85 | 5 | 97.96 | 0.85 | 5 |
| 3 | 1.01 | 0.9 | 5 | 99 | 0.9 | 5 |
| 4 | 1.04 | 0.85 | 5 | 98.94 | 0.85 | 5 |
| 5 | 0.53 | 0.9 | 5 | 99.46 | 0.9 | 5 |

The optimal MMD threshold for the CASIA dataset, recorded across all mean EER and mean Accuracy results from **Table 1**, is 10 pixels, suggesting some rotational variation still persists in the CASIA dataset after the pre-alignment. This variance is likely to occur away from the valley points of the palms towards the wrists. The lowest EER and highest Accuracy results (**Table 2**) are recorded when the MMD threshold lies within the 5-15 range, except when the template size is three, where the lowest EER is recorded at an MMD threshold of 50.

**Table 4**: Lowest EER % and highest Accuracy % values recorded from any combination from PolyU.

| Template Size | Lowest EER % | RT Val. | MMD Tr. | Highest ACC % | RT Val. | MMD Tr. |
|---|---|---|---|---|---|---|
| 1 | 2.84 | 0.85 | 5 | 96.3 | 0.85 | 10 |
| 2 | 1.4 | 0.85 | 5 | 98.45 | 0.85 | 5 |
| 3 | 0.79 | 0.9 | 5 | 99.25 | 0.9 | 5 |
| 4 | 0.41 | 0.85 | 5 | 99.42 | 0.85 | 5 |
| 5 | 0.21 | 0.9 | 5 | 99.78 | 0.9 | 5 |

**Table 5**: Lowest mean EER % and the highest mean Accuracy % values on the PUT dataset.

| Template Size | Mean EER % | RT Val. | MMD Tr. | Mean ACC % | RT Val. | MMD Tr. |
|---|---|---|---|---|---|---|
| 1 | 1.51 | 0.85 | 40 | 98.35 | 0.85 | 40 |
| 2 | 0.45 | 0.9 | 30 | 99.64 | 0.9 | 30 |
| 3 | 0.16 | 0.9 | 20 | 99.9 | 0.85 | 15 |
| 4 | 0.12 | 0.9 | 30 | 99.94 | 0.8 | 20 |
| 5 | 0.02 | 0.9 | 20 | 99.98 | 0.85 | 20 |

**Table 6**: Lowest EER % and highest Accuracy % values from PUT.

| Template Size | Lowest EER % | RT Val. | MMD Tr. | Highest ACC % | RT Val. | MMD Tr. |
|---|---|---|---|---|---|---|
| 1 | 0.97 | 0.85 | 40 | 98.95 | 0.9 | 30 |
| 2 | 0.05 | 0.85 | 15 | 99.94 | 0.85 | 15 |
| 3 | 0 | 0.75 | 10 | 100 | 0.75 | 10 |
| 4 | 0 | 0.7 | 5 | 100 | 0.7 | 5 |
| 5 | 0 | 0.65 | 5 | 100 | 0.65 | 5 |

The results obtained using all the gallery-probe combinations on the CASIA dataset indicate that the lowest EER values, which utilise 100% of the CASIA dataset, are higher than those presented in our previous study [9]. In that study, the full palmar region was used for the ROI at a 60% scale, and only 80% of the dataset was employed to verify the performance of the MMD filter. Additionally, when scaling the ROI to 128 × 128 pixels, the initial cropped ROI area of some palm images was smaller than this size and was scaled up, potentially amplifying compression artefacts.

On **Table 3**, MMD threshold of 5 pixels was consistently recorded except for mean Accuracy when using a template size of one with the PolyU dataset.

This consistency was further evidenced by the fact that the lowest EER and highest Accuracy recorded from any single gallery-probe combination for all the template sizes (**Table 4**) were observed at the corresponding mean RT+MMD combination. The low MMD threshold confirms the presence of only minor rotational variances within the PolyU dataset, indicating a high degree of uniformity in the orientation of the images and features being analysed.

On the PUT dataset (**Table 5**), MMD thresholds are initially high when using a lower template size but gradually decrease to a moderate value of 20 as the template size increases. These higher MMD thresholds in the PUT dataset result from its acquisition conditions, as it was captured without a hand placement guide, and rotational alignment was not normalized. Consequently, greater rotational variations are present, requiring higher MMD thresholds to compensate for these inconsistencies.

**Table 6** shows the lowest EER and highest Accuracy from any single combination across the system on the PUT dataset. When these values reached their extremes at the specified RT and MMD combinations, we observed that they remained at these extremes despite increases in RT and MMD values.

### 6.2 MMD RT Recommendations for Generalisation

The following guidelines generalise findings for application to other datasets and real-world biometric systems. Both MMD thresholds and RT values should be adapted based on dataset characteristics, specifically acquisition setup and variation levels.

For contact-based datasets with hand placement guides (e.g., PolyU), palm-vein variation is minimal, warranting very low MMD thresholds. For datasets with free hand presentation and visible finger valleys where rotational normalisation is feasible (e.g., CASIA), lower to medium thresholds suffice. For datasets lacking visible finger valleys where rotational normalisation fails (e.g., PUT), even in contact settings without placement guidance, higher MMD thresholds accommodate greater rotational variation. Larger ROI scales may require further MMD threshold increases. Note: upscaling palm-vein images risks introducing artefacts and degrading feature quality.

Lowe's [14] original suggestion of 0.7–0.8 proved suboptimal across all three datasets. Higher RT values consistently delivered superior results. This improvement stems from palm-vein images' confined feature space and low contrast, which constrain feature distinctiveness compared to conventionally contrasted imagery. Consequently, an RT value between 0.85–0.9 is recommended for palm-vein recognition systems.

### 6.3 Performance Comparison Against Other Systems

The most commonly reported performance metric in existing palm-vein recognition studies is the EER, followed by Accuracy. These studies do not employ a uniform experimental setup and differ in various aspects, including database size, template size, the division of training and test data, and the selection of subsets between left and right-hand images. To enable a meaningful comparison, studies were selected that utilised an experimental setup most similar to the one employed in this research.

Caballero et al. [55] explored various LBP-based methods and parameters, comparing the performance of LBP and LBPU using the chi-square ($\chi^2$) distance. They utilised all spectral subsets from the CASIA dataset, including the NIR bands at 850 nm and 940 nm, and reported EER. The ROI was obtained by cropping the central palmar area of rotation-normalised palm images, with HE applied to enhance contrast.

Nayar et al. [56] introduced a graph-based approach that utilises enhanced maximum curvature to extract vein patterns, representing them as nodes and edges to generate a user-specific key and a cryptographically secure pseudo-random number generator. They utilised three datasets: CASIA, PUT, and VERA [57]. The ROI was cropped to the central palmar area of rotation-normalised images in the CASIA and VERA datasets. CLAHE and a median filter were used for contrast enhancement. They reported EER and RR.

Qin et al. [58] employed Label Enhancement-Based Multiscale Vein Transformer (LE-MSVT) and adversarial learning for vein pattern recognition and data augmentation. They utilised three datasets: PolyU, VERA, and Tongji [59], with a 50%-50% split for training and testing. They reported closed-set Accuracy and EER, as well as open-set Accuracy results. No contrast enhancement method was discussed.

Babalola et al. [60] combined texture-based features extracted using BSIF with a CNN model, employing a decision-level fusion strategy. They utilised five datasets: CASIA, PUT, VERA, Tongji, and FYO [61], with a cropped ROI where necessary. They only reported Accuracy results.

**Table 7** presents a comparison of our results against those from existing studies. It is evident upon comparison that the proposed method generally outperforms existing methods in EER and Accuracy across all datasets. As the comparator studies report results from a single train/test split rather than an average across gallery-probe combinations, the lowest EER and highest Accuracy values recorded from a single combination in this study provide the appropriate basis for comparison rather than the mean.

**Table 7**: EER and Accuracy results comparison with existing work.

| Performance Metric \| Dataset | Performance reported | Experimental setup |
|---|---|---|
| EER % \| CASIA | LBP + $\chi^2$ distance [55] \| Template size – 3<br>9.5 \| 850nm<br>7.33 \| 940nm | *1200 images – Full subsets consisting of 200 palms on each wavelength.* |
| | Proposed \| Template size – 3<br>1.081 – Mean<br>**0.36** – Lowest | *1200 images – Full 850nm subset consisting of 200 palms.* |
| | Graph-based [56] \| Template size – 4<br>0.02 | *600 images – Left and right-hand images from 50 subjects in the 940nm subset, consisting of 100 palms.* |
| | Proposed \| Template size – 4<br>0.63 – Mean<br>**0.0** – Lowest | *1200 images – Full 850nm subset consisting of 200 palms.* |
| EER % \| PUT | Graph-based [56] \| Template size – 4<br>**0.0** | *1200 images – Full 880nm palm-vein subset consisting of 100 palms.* |
| | Proposed \| Template size – 4<br>0.12 – Mean<br>**0.0** – Lowest | *1200 images – Full 880nm palm-vein subset consisting of 100 palms.* |
| EER % \| PolyU | LE-MSVT [58] \| Template size – 5<br>**0.26** | *Training and test split is 50% / 50% of the full 880nm subset, consisting of 500 palms. EER results are reported using a template size of 5.* |
| | Proposed \| Template size – 5<br>0.53 – Mean<br>0.21 – Lowest | *6000 images - Full 880nm subset consisting of 500 palms.* |
| Accuracy % \| CASIA | BSIF + CNN [60] \| Template size ~5<br>99.83 | *Training and test split is 80% / 20% of the full dataset with 1200 images, consisting of 200 palms. This approximates to a template size of 5. NIR wavelength not specified.* |
| | Proposed \| Template size – 5<br>99.52 – Mean<br>**100** – Highest | *1200 images – Full 850nm subset consisting of 200 palms.* |
| Accuracy % \| PUT | BSIF + CNN [60] \| Template size ~10<br>99.0 | *Training and test split is 80% / 20% of the full 880nm palm-vein subset with 1200 images, consisting of 100 palms. This approximates to a template size of 10.* |
| | Proposed \| Template size – 5<br>99.98 – Mean<br>**100** – Highest | *1200 images – Full 880nm palm-vein subset consisting of 100 palms.* |
| Accuracy % \| PolyU | LE-MSVT [58] \| Template size – 5<br>99.34 | *Training and test split is 50% / 50% of the full 880nm subset, consisting of 500 palms. Accuracy results are reported using template sizes of 4, 5, 6.* |
| | Proposed \| Template size – 5<br>99.46 – Mean<br>**99.78** – Highest | *6000 images - Full 880nm subset consisting of 500 palms.* |

## 7. Conclusions

This study introduced ILACS-BGOT, a contrast enhancement method designed to address limitations in existing palm-vein imaging techniques such as the loss of fine feature details and the blocky artefacts caused by tile-based methods. By improving upon ILACS-LGOT, ILACS-BGOT enables better scalability to larger tile sizes while preserving critical biometric features.

Extensive evaluation across three benchmark datasets (CASIA, PolyU, and PUT) demonstrated the method's effectiveness, with consistently low EER and high Accuracy rates. These results confirm that ILACS-BGOT enhances the discriminative quality of palm-vein features.

Beyond the empirical performance, this work contributes a generalisable enhancement framework which can potentially be extended beyond palm-vein recognition to other related biometric modalities such as finger vein and palmprint recognition. Its potential extends further to general low-contrast image enhancement tasks in broader computer vision domains.

Future research could explore real-time implementations, robustness in unconstrained environments, or integration into deep learning pipelines for end-to-end biometric systems. The open-sourcing of ILACS-BGOT aims to support reproducibility and foster further innovation in biometric imaging.

The code for the ILACS-BGOT method is available at: [https://github.com/kaveenperera/ILACS-Enhancement](https://github.com/kaveenperera/ILACS-Enhancement) under Mozilla Public License Version 2.0.